\documentclass[sigconf, nonacm]{acmart}
\usepackage{multirow} 
\usepackage{xcolor}
\usepackage[most]{tcolorbox} 
\usepackage{microtype}
\definecolor{deltagreen}{RGB}{235, 255, 235}
\definecolor{textgreen}{RGB}{0, 150, 0}
\definecolor{textred}{RGB}{200, 0, 0}
\usepackage[table]{xcolor} 
\usepackage[table]{xcolor} 
\usepackage{graphicx}
\usepackage{subcaption}
\definecolor{deltagreen}{rgb}{0.8, 0.98, 0.8}
\definecolor{refreshinggreen}{rgb}{0.92, 1, 0.94}
\tcbuselibrary{skins}
\usepackage{graphicx}

\acmISBN{978-1-4503-XXXX-X/2018/06}

\begin{document}

\title{A Progressive Visual-Logic-Aligned Framework for Ride-Hailing Adjudication}


\author{Weiming Wu}
\authornote{Equal contribution. Work done during internship at Didichuxing Co. Ltd.} 
\email{wuwm23@smail.nju.edu.cn}
\affiliation{%
  \institution{Nanjing University}
  \city{}
  \state{}
  \country{}
}

\author{Zi-Jian Cheng}
\authornotemark[1] 
\email{chengzj@lamda.nju.edu.cn}
\affiliation{%
  \institution{Nanjing University}
  \city{}
  \state{}
  \country{}
}

\author{Jie Meng}
\email{jmengjie@didiglobal.com}
\affiliation{%
  \institution{Didichuxing Co. Ltd}
  \city{}
  \state{}
  \country{}
}

\author{Peng Zhen}
\email{zhenpeng@didiglobal.com}
\affiliation{%
  \institution{Didichuxing Co. Ltd}
  \city{}
  \state{}
  \country{}
}

\author{Shan Huang}
\email{lattehuang@didiglobal.com}
\affiliation{%
  \institution{Didichuxing Co. Ltd}
  \city{}
  \state{}
  \country{}
}

\author{Qun Li}
\email{liquntracy@didiglobal.com}
\affiliation{%
  \institution{Didichuxing Co. Ltd}
  \city{}
  \state{}
  \country{}
}

\author{Guobin Wu}
\email{wuguobin@didiglobal.com}
\affiliation{%
  \institution{Didichuxing Co. Ltd}
  \city{}
  \state{}
  \country{}
}

\author{Lan-Zhe Guo}
\authornote{Corresponding author.} 
\email{guolz@lamda.nju.edu.cn}
\affiliation{%
  \institution{Nanjing University}
  \city{}
  \state{}
  \country{}
}
\renewcommand{\shortauthors}{Wu et al.}

\begin{abstract}
The efficient adjudication of responsibility disputes is pivotal for maintaining marketplace fairness. However, the exponential surge in ride-hailing volume renders manual review intractable, while conventional automated methods lack the reasoning transparency required for quasi-judicial decisions. Although Multimodal LLMs offer a promising paradigm, they fundamentally struggle to bridge the gap between general visual semantics and rigorous evidentiary protocols, often leading to perceptual hallucinations and logical looseness. To address these systemic misalignments, we introduce \textbf{RideJudge}, a Progressive Visual-Logic-Aligned Framework. Instead of relying on generic pre-training, we bridge the semantic gap via SynTraj, a synthesis engine that grounds abstract liability concepts into concrete trajectory patterns. To resolve the conflict between massive regulation volume and limited context windows, we propose an Adaptive Context Optimization strategy that distills expert knowledge, coupled with a Chain-of-Adjudication mechanism to enforce active evidentiary inquiry. Furthermore, addressing the inadequacy of sparse binary feedback for complex liability assessment, we implement a novel Ordinal-Sensitive Reinforcement Learning mechanism that calibrates decision boundaries against hierarchical severity. Extensive experiments show that our RideJudge-8B achieves 88.41\% accuracy, surpassing 32B-scale baselines and establishing a new standard for interpretable adjudication.
\end{abstract}



 \maketitle

\section{Introduction}
 The rapid development of on-demand ride-hailing platforms has fundamentally revolutionized the landscape  of urban mobility and transportation dynamics. Within this ecosystem, the adjudication of responsibility disputes serves as a cornerstone for maintaining fairness and marketplace stability. As these transportation systems increasingly rely on decentralized operational models, the volume of service disputes, ranging from route deviations to cancellation disagreements, has grown exponentially. Consequently, efficiently and objectively resolving these disputes has emerged as a vital challenge in the field of intelligent operational systems~\cite{zhang2011detecting, liu2020online}.

\textbf{Ride-hailing adjudication} is a complex task that necessitates the integration of heterogeneous data sources, such as vehicle trajectories, passenger-driver behavioral records, and order metadata, to perform reasoning according to platform rules. Conventionally, automated solutions for this task have relied on traditional models~\cite{xgboost, lightgbm}, which often suffer from limited accuracy when processing complex multimodal interactions. More critically, these methods typically yield only adjudication labels, lacking the capacity to provide the reasoning trails required to justify verdicts. 

Recently, Multimodal Large Language Models (MLLMs)~\cite{gpt4v, gemini15, llama3herd, qwen3vl} have emerged as a promising paradigm for automated liability adjudication. By leveraging their generative capabilities, these models theoretically enable the synthesis of heterogeneous data sources to derive interpretable verdicts. However, the direct application of off-the-shelf MLLMs to this specialized domain is hindered by intrinsic misalignments between general pre-training objectives and the rigorous demands of judicial reasoning.

\textbf{Why General MLLMs Fail?}  We identify three disconnects that prevent general models from functioning as reliable adjudicators:

\textbf{General Perceptual vs. Domain-Specific Grounding.} Existing vision encoders, predominantly pre-trained on natural image-caption pairs, exhibit a semantic rupture when interpreting abstract, schematic navigation maps. While these models can recognize basic primitives, they lack the specialized perceptual grounding required to translate visual anomalies, such as subtle trajectory drift or abnormal stops, into precise juridical liability concepts. This misalignment renders generic models incapable of distinguishing fine-grained violation patterns from standard driving fluctuations.

\textbf{Probabilistic Generation vs. Deterministic Logical Deduction.} Adjudication requires a rigorous reasoning schema involving identifying facts, selecting applicable rules, and deducing verdicts, which stands in contrast to the probabilistic nature of LLM generation. Despite their promise in open-ended tasks, MLLMs often lack the logical robustness necessary for multi-hop, evidentiary reasoning. In high-stakes scenarios, they struggle to maintain a rigorous evidentiary chain, frequently yielding verdicts that are either logically decoupled from established facts or internally inconsistent with the cited platform regulations

\textbf{Fixed Contextual Capacity vs. Open-World Rule Scaling.} Unlike standard tasks with self-contained contexts, ride-hailing adjudication operates within a dynamic and expansive knowledge ecosystem comprising evolving regulations and historical precedents. This legal corpus often exceeds the finite context windows of standard MLLMs. Moreover, general models lack the scenario-aware retrieval mechanisms needed to filter relevant statutes from massive repositories, leading to the injection of contextual noise and a degradation in the precision of the reasoning process.

To address these challenges, we propose a \textbf{Progressive Visual-Logic-Aligned Framework} for the ride-hailing adjudication task. Our core insight is to treat adjudication as a cognitive mirroring process, where the model learns to bridge the epistemic gap between raw spatiotemporal dynamics and high-level juridical reasoning, simulating the rigorous workflow of human experts.

To bridge the perception and reasoning gaps, we first introduce an automated data synthesis engine. This method includes a visual perception synthesis module, SynTraj, which employs programmatic simulation to generate fine-grained trajectory data rich in specific violation semantics. Additionally, it incorporates a logic synthesis modulehat utilizes a chain-of-adjudication reasoning process to construct rigorous reasoning trails with evidence that mirror expert decision-making patterns.

To resolve the utilization bottleneck of massive knowledge, we introduce an adaptive context optimization strategy. To handle the scale of rules and cases, this module employs a scenario-aware rule pruning mechanism to strictly filter relevant regulations and a dynamic case retrieval system to extract expert guidance from historical precedents, which allows the model to consult external knowledge and align with domain protocols without context overload.

Finally, addressing the limitations of traditional GRPO answer reward where sparse binary feedback fails to distinguish error severity, we design an Ordinal-Sensitive Reward. By assigning graded signals based on the semantic proximity between predictions and ground truths, this mechanism mitigates reward sparsity. Capitalizing on this, we implement a three-stage progressive training paradigm to align the model from visual semantic understanding to complex, hierarchically consistent decision-making.

Our contributions can be summarized as follows: 1) We propose a specialized multimodal framework for ride-hailing adjudication that effectively aligns visual trajectory data with rigorous juridical rules; 2) We design an integrated solution featuring automated visual-logic synthesis, adaptive context optimization for knowledge integration, and a progressive training paradigm with an ordinal-sensitive reward; and 3) We validate the effectiveness and reliability of our method through extensive experiments on multiple real-world datasets, demonstrating superior performance in complex adjudication tasks.

\section{Preliminaries}

In this section, we formalize the operational logic of the ride-hailing platform, define the comprehensive data structures for orders, and formulate the core task of intelligent adjudication.

\subsection{Ride-Hailing Order} 

We first establish the formal definitions for the data structures generated during the ride-hailing service lifecycle.

\textbf{Definition 1: Ride-Hailing Order.}
A ride-hailing order is composed of textual metadata and visual spatial data.

\textit{1) Textual Information ($O_{text}$):}
Upon order acceptance, the system generates an initial static information tuple:
\begin{equation}
o_{init} = (l_{driver}, l_{start}, l_{end}, P_{driver}, P_{pass})
\end{equation}
where $l_{driver}$, $l_{start}$, and $l_{end}$ denote the driver's acceptance point, the order start point, and the destination; $P_{driver}$ and $P_{pass}$ represent the generic profiles of the driver and passenger, respectively.

As the order progresses from the pickup to its final termination, resulting from either successful completion or abnormal cancellation, the system records dynamic behavioral features. We define the accumulated textual features as:
\begin{equation}
    O_{text} = \{o_{init}, \mathcal{F}_{driver}, \mathcal{F}_{pass}\}
\end{equation}
Here, $\mathcal{F}_{driver} = \{d_1, d_2, \dots, d_i\}$ represents accumulated driver behavioral statistics such as stationary duration and detour distance; $\mathcal{F}_{pass}$ represents the  passenger behavioral statistics.

\textit{2) Visual Information $O_{image}$:}
To effectively capture the spatial semantics and road network structure, we adopt a rasterization-based approach inspired by recent advances in visual trajectory modeling~\cite{zhang2016deep, bojarski2016end, ma2019learning}. Specifically, we construct a visual representation $O_{image}$ by rendering both the driver's executed path $\mathcal{T}_{real}$ and the algorithmically planned route $\mathcal{T}_{nav}$ onto a roadmap that contains the detailed road layout. This spatial superposition serves as a strong visual prior and enables the model to intuitively capture fine-grained behavioral patterns of both the driver and the passenger, such as unexpected detours or abnormal stops~\cite{liu2020online}, which are critical for liability assessment.

Finally, the comprehensive Multimodal Context is defined as $\mathcal{O} = \{O_{text}, O_{image}\}$.

\textbf{Definition 2: Disputed Order.}
A \textit{Disputed Order} is defined as an order instance where the trip is cancelled by either the driver or the passenger abnormally. This cancellation event marks the termination of the service recording and flags the order for potential liability review.

\subsection{The Ride-Hailing Adjudication Task}

To maintain the ecological balance of the marketplace and ensure fairness~\cite{ zha2021fairness}, the platform must determine liability for every disputed order. This process requires the platform to arbitrate the responsibility based on explicit regulations, categorizing the driver and passenger behaviors into a hierarchical label space $\mathcal{Y}$ which consists of multiple levels representing different degrees of liability.

Given the comprehensive order context $\mathcal{O}$ and the external Adjudication Knowledge Base $\mathcal{K}$, our goal is to construct an intelligent model that functions as a mapping $F$:
\begin{equation}
 y = F(\mathcal{O}, \mathcal{K})
\end{equation}
Here, $y \in \mathcal{Y}$ is the predicted liability verdict. The model must effectively integrate the heterogeneous information in $\mathcal{O}$ with specific rules in $\mathcal{K}$ to derive a logically sound verdict.

Conventionally, automated solutions for this task have relied on traditional discriminative models~\cite{xgboost, lightgbm}. However, these approaches often suffer from limited accuracy when processing complex multimodal spatial-temporal data. More critically, these methods typically yield only classification labels, lacking the capacity to provide the transparent reasoning trails required to justify verdicts~\cite{rudin2019stop}, and are prone to robustness issues when facing long-tail scenarios.

\begin{figure*}[t]
    \centering
    \includegraphics[width=1\textwidth]{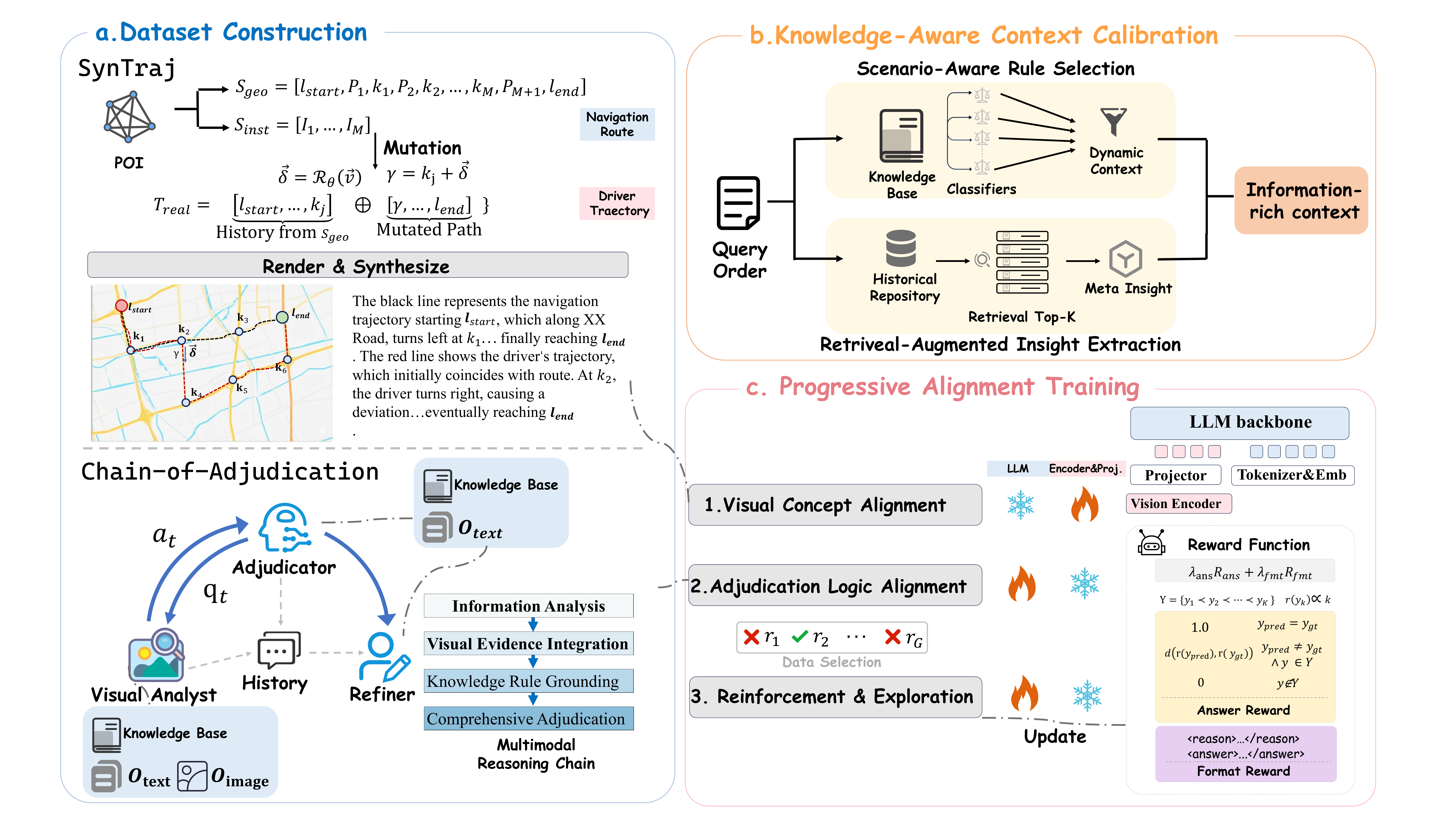}
    \caption{
    The pipeline consists of three pivotal phases: 
    (1) \textbf{Automated Data Synthesis} (Sec.~\ref{sec:data_synthesis}), which bridges domain gaps via two specialized modules: \textit{SynTraj Construction} for visual-linguistic alignment and \textit{Chain-of-Adjudication Synthesis} for logical reasoning reconstruction; 
    (2) \textbf{Knowledge-Aware Context Refinement} (Sec.~\ref{sec:knowledge_refinement}), capable of dynamic rule pruning and expert precedent extraction; 
    and (3) \textbf{Progressive Juridical Alignment} (Sec.~\ref{sec:progressive_alignment}), a multi-stage training paradigm culminating in OS-rewarded reinforcement learning for precise decision boundary alignment.}
    \label{fig:main_pipeline}
    \vspace{-4mm}
\end{figure*}

To address these challenges and automate the workflow, we propose a Progressive Visual-Logic-Aligned Framework. As illustrated in Figure \ref{fig:main_pipeline}, our method consists of three components designed to bridge the gap between multimodal data and expert adjudication logic. These components include an Automated Data Synthesis Framework for bridging domain gaps via visual-linguistic alignment and logical reasoning reconstruction~\cite{wei2022chain}, an Adaptive Context Optimization Strategy for dynamic rule pruning and expert precedent extraction, and a Progressive Training Paradigm that aligns the model from semantic understanding to complex decision-making via ordinal-sensitive reinforcement learning.
hai
\section{Training Data Construction}
\label{sec:data_synthesis}

\subsection{SynTraj: Bridging Visual Semantics}
\label{sec:syntraj_construction}

To establish a fundamental alignment between ride-hailing concepts and their visual representations, we propose an automated \textbf{Syn}thetic \textbf{Traj}ectory generation framework. SynTraj injects spatiotemporal priors into the model, ensuring it can perceptually ground abstract adjudication behaviors into concrete trajectory patterns. The pipeline consists of two tightly coupled processes named structural route planning and behavioral trajectory simulation.

\textbf{Navigation Route Planning.}
We first establish the ground truth by constructing valid navigation routes. We sample Point of Interest pairs from major cities in China and query standard navigation APIs\footnote{In this work, we utilize the open platform: \url{https://lbs.amap.com}.} to obtain route data. Drawing upon standard notations in trajectory mining literature, we decompose the route into a synchronized geometric backbone and an instruction sequence. The Geometric Sequence $S_{geo}$ is modeled as a continuous coordinate chain linking the start point $l_{start}$ and end point $l_{end}$, interspersed with critical intersection nodes $k$ and dense segment points $P$:
\begin{equation}
    S_{geo} = [l_{start}, P_{1}, k_{1}, P_{2}, k_{2}, \dots, k_{M}, P_{M+1}, l_{end}]
\end{equation}
where $k_{m}$ represents the $m$-th critical intersection node and $P_{m}$ denotes the dense sequence of GPS coordinates constituting the road segment between $k_{m-1}$ and $k_{m}$. Synchronized with this is the Instruction Sequence $S_{inst}$ containing $M$ instructions. Each instruction $I_m$ specifies the driving behavior along segment $P_{m}$ and the required maneuver at node $k_{m}$.

\textbf{Driver Trajectory Synthesis.}
Based on the planned route, we simulate diverse driver behaviors ranging from strict compliance to specific violations. For compliant behaviors, we simulate a driver faithfully following the platform guidance. We generate a trajectory that traverses $S_{geo}$ by applying Gaussian noise perturbations to the coordinates to simulate intrinsic GPS errors and natural driving fluctuations. This ensures the trajectory remains structurally aligned while exhibiting realistic sensor irregularities.

For abnormal behaviors, our engine supports multiple mutation strategies (detailed comprehensively in Appendix\ref{app:mutation_protocols}). Here, we illustrate the mechanism using the \textit{"Unintentional Deviation"} scenario as a representative example. We first randomly select an intersection node $k_j$ from $S_{geo}$ as the anchor. To simulate the deviation dynamics, we define the intended direction vector $\vec{v} = p_{next} - k_j$, where $p_{next}$ denotes the first point immediately following $k_j$ in the subsequent segment. We then apply a rotation operator $\mathcal{R}_\theta$ to generate a deviation vector $\vec{\delta} = \mathcal{R}_\theta(\vec{v})$ with a randomized magnitude. The mutated anchor point is calculated as $\gamma = k_j + \vec{\delta}$. We subsequently query the API for a new path from $\gamma$ to the destination and stitch this deviation path with the historical compliant segment. This process yields a final trajectory $T_{real}$ that maintains geometric consistency while injecting precise liability semantics:
\begin{equation}
    T_{real} = \underbrace{[l_{start}, \dots, k_j]}_{\text{History from } S_{geo}} \oplus \underbrace{[\gamma, \dots, l_{end}]}_{\text{Mutated Path}}
\end{equation}

\textbf{Multimodal Pair Generation.}
Finally, we generate the aligned visual-textual pairs. For the visual input, we employ the protocol defined in the preliminary section to render the synthesized driver trajectory $T_{real}$ and the original navigation route $S_{geo}$ into an image. This ensures the visual difference explicitly reflects the behavioral patterns. For the textual label, we utilize the meta-information from the instruction sequence $S_{inst}$ to instantiate natural language descriptions. This process yields a dataset of 12,585 high-quality image-caption pairs, enabling the model to learn the correspondence between visual trajectory patterns and textual adjudication concepts without the noise inherent in real-world data.
\subsection{Chain-of-Adjudication: Synthesizing Adjudication Reasoning}
\label{sec:logic_alignment}

To transcend the limitations of existing black box models, which often decouple reasoning from raw evidence~\cite{rudin2019stop, lipton2018mythos, doshi2017towards}, we propose \textbf{Chain-of-Adjudication} (CoA), which synthesizes rigorous and evidence-backed reasoning chains mirroring professional judicial workflows. Unlike standard end-to-end approaches~\cite{llava, capsfusion} that indiscriminately process all modalities, CoA is designed to simulate the procedural rigor of a human judge, inspired by recent advances in multimodal reasoning~\cite{wei2022chain, visionr1, vlmr1}.

It begins from the \textbf{Adjudicator}, which is an LLM serving as the reasoning engine, to assimilating the textual context $O_{text}$ while concurrently retrieving specific liability clauses from $\mathcal{K}$. Crucially, Adjudication is intentionally isolated from direct visual input. This structural constraint forces the agent to transition from passive perception to active inquiry~\cite{proreason, jian2024large}, meaning that instead of hallucinating visual details from captions, Adjudication must formulate precise and hypothesis-driven queries to verify the factual predicates required by the retrieved adjudication rules.

To satisfy these evidentiary inquiries, the framework incorporates a \textbf{Visual Analyst} functioning as an auxiliary perceptual anchor. Analyst receives specific verification queries from Adjudicator and examines $O_{image}$ to provide objective fact-based descriptions. This establishes a robust Iterative Verification Loop wherein the Analyst hypothesizes potential violations based on the rules and the Analyst validates or refutes them with trajectory evidence. This multi-turn interaction ensures that every logical step is explicitly anchored in spatiotemporal reality, effectively bridging the gap between abstract liability concepts and concrete visual patterns.

The raw interaction history typically retains conversational redundancies that are suboptimal for training. To address this, we deploy a \textbf{Reasoning Refiner} acting as a meta-cognitive editor to distill the fragmented dialogue into a coherent adjudication path. Specifically, Refiner restructures the content into a standardized four-stage comprising: 1) \textit{Information Analysis}, which systematically summarizes the order metadata and dispute context; 2) \textit{Visual Evidence Integration}, which incorporates the objective trajectory facts verified by Analyst; 3) \textit{Rule Grounding}, which maps the established facts to specific liability clauses within ($\mathcal{K}$); and 4) \textit{Comprehensive Adjudication}, which performs the final logical deduction to derive the verdict.

Finally, we implement a rigorous \textbf{Data Selection} process to guarantee corpus reliability. We systematically filter out instances where the synthesized verdict $\hat{y}$ diverges from the human-annotated $y_{gt}$ or where the historical order is marked as Ambiguous. This stringent filtering yields a final corpus of 14,582 samples, ensuring that the dataset consists exclusively of high-confidence and evidence-supported reasoning trajectories.
\section{Knowledge-Aware Context Refinement}
\label{sec:knowledge_refinement}

Ride-hailing adjudication requires consulting a massive Knowledge Base ($\mathcal{K}$) and referencing extensive historical precedents. However, directly feeding this extensive corpus into the model exceeds the context window limits of standard MLLMs and introduces irrelevant noise, which distorts the reasoning process. To address this, we propose an Adaptive Context Optimization strategy, inspired by recent advances in agentic context engineering~\cite{agentic_context} and optimization-based reasoning~\cite{deepseekr1}. This approach selectively filters relevant rules and distills expert consensus from historical data to construct a precise and information-rich context for the final reasoning stage.

\subsection{Scenario-Aware Rule Calibration}
\label{sec:rule_calibration}

To efficiently identify applicable statutes from the fine-grained Rule Base $\mathcal{K} = \{r_1, \dots, r_N\}$, we propose a Decomposed Ensemble Calibrator.

\textbf{Filter Training.}
We formulate the rule calibration task as a multi-label classification problem via a binary decomposition strategy~\cite{tsoumakas2007multi}. We first construct a specialized dataset $\mathcal{D}_{cal}$ consisting of representative order samples annotated with binary applicability vectors. We decompose the multi-label task into an ensemble of $N$ independent binary classifiers $\mathcal{E} = \{f_1, \dots, f_N\}$. Each classifier $f_i$, implemented via scalable tree boosting models~\cite{xgboost}, determines the binary relevance of a specific rule $r_i$ based on a generic input order $O$:
\begin{equation}
    f_i(O) = 
    \begin{cases} 
    1 & \text{if rule } r_i \text{ is applicable} \\
    0 & \text{otherwise}
    \end{cases}
\end{equation}
This decomposition ensures robustness against statistical dependencies between distinct regulations and allows for the flexible use of various classifiers.

\textbf{Application.}
During the inference phase, given a specific query order $O^{query}$, we employ this trained ensemble to prune the knowledge base. We aggregate the outputs to construct a scenario-specific rule subset $\mathcal{K}'$ by retaining only the rules predicted as positive:
\begin{equation}
    \mathcal{K}' = \{ r_i \in \mathcal{K} \mid f_i(O^{query}) = 1 \}
\end{equation}
By strictly filtering out irrelevant clauses, this mechanism significantly reduces context redundancy and prevents the reasoning model from generating hallucinations based on inapplicable regulations.

\subsection{Retrieval-Augmented Insight Extraction}
\label{sec:insight_extraction}

To leverage tacit adjudication knowledge, we implement a Retrieve-then-Extract paradigm, integrating Retrieval-Augmented Generation~\cite{lewis2020rag} with Case-Based Reasoning.

\textbf{Vectorized Precedent Retrieval.}
We maintain a dynamically updated historical repository $\mathcal{D}_{hist} = \{(O^{(j)}_{text}, y^{(j)})\}_{j=1}^{N}$. Crucially, to prevent data leakage, this repository consists strictly of orders processed prior to the timestamp of the current query. Given a query $O^{query}_{text}$, we employ a dense retriever to identify the Top-$K$ semantic neighbors $\mathcal{N}_K$ based on vector similarity:
\begin{equation}
    \mathcal{N}_K = \mathop{\arg\max}_{\mathcal{S} \subset \mathcal{D}_{hist}, |\mathcal{S}|=K} \sum_{j \in \mathcal{S}} \frac{\mathbf{E}(O^{query}_{text}) \cdot \mathbf{E}(O^{(j)}_{text})}{\|\mathbf{E}(O^{query}_{text})\|_2 \|\mathbf{E}(O^{(j)}_{text})\|_2}
\end{equation}
where $\mathbf{E}(\cdot)$ denotes the embedding function, and $\|\cdot\|_2$ represents the $L_2$ norm.

\textbf{Meta-Insight Abstraction.}
To avoid information sparsity arising from raw text concatenation, we employ a Summary Agent powered by an LLM. Drawing inspiration from verbal reinforcement learning mechanisms~\cite{shinn2024reflexion}, this agent analyzes the retrieved cohort $\mathcal{N}_K$ to identify statistical commonalities and adjudication patterns. It outputs a concise Meta-Insight $I_{syn}$ via a dedicated summarization LLM, denoted as $g$:
\begin{equation}
    I_{syn} = g(\mathcal{N}_K)
\end{equation}
This synthesized insight bridges the gap between raw history and current decision-making, providing the model with expert references that stabilize the reasoning process in complex scenarios.

\section{Progressive Alignment Framework}
\label{sec:progressive_alignment}

To bridge the gap between general multimodal capabilities and the rigorous demands of adjudication, we propose a three-stage progressive training framework.

\textbf{Stage 1: Visual Concept Alignment.} 
To establish the fundamental alignment between ride-hailing concepts and visual representations, we employ the SynTraj dataset. In this stage, we freeze the language model while optimizing the vision encoder and projector. This process enables the model to perceptually ground abstract adjudication rules into concrete trajectory patterns without altering the pre-trained knowledge base.

\textbf{Stage 2: Adjudication Logic Alignment.} 
We subsequently conduct Supervised Fine-Tuning on multimodal reasoning dataset synthesized via the Chain-of-Adjudication framework. We finetune the LLM backbone to maximize the likelihood of the expert reasoning chain. This stage  aligns the reasoning process of the model with professional adjudication protocols and strengthens its instruction following capabilities.

\textbf{Stage 3: Reinforcement and Exploration.}
In the final stage, we employ Reinforcement Learning to further enhance the model's robustness and explore the decision boundaries of complex cases.

\textbf{Divergence-Aware Data Selection.}
To maximize reinforcement learning efficiency, we implement a difficulty-aware selection strategy. We assess the complexity of each instance by performing $N=10$ stochastic rollouts using the Stage 2 model. This process generates a set of correctness scores where $1$ represents a correct verdict and $0$ represents an incorrect one. We calculate the average consistency score $S_{avg}$ and strictly retain instances that satisfy the criterion $0.2 \le S_{avg} \le 0.8$. This filtering mechanism effectively eliminates both trivial and intractable cases to ensure the model focuses on samples where it currently exhibits reasoning ambiguity. Consequently, we retain approximately 2,000 filtered samples for the subsequent training phase.

\textbf{Ordinal-Sensitive Reward}
We utilize the DAPO algorithm~\cite{yu2025dapo} for optimization. To address the limitations of sparse binary feedback in adjudication, we propose a novel Ordinal-Sensitive Reward mechanism. Deviating from traditional methods that treat all errors equally, we assign graded reward signals based on the semantic proximity between predictions and ground truths. This approach effectively mitigates reward sparsity and guides the model to distinguish between error level, ensuring precise and hierarchically consistent decision-making.

We formalize the liability label space as an ordered set $\mathcal{Y} = \{y_1 < y_2 < \dots < y_K\}$ where the rank mapping $r(y_k)$ corresponds to the ordinal index $k$. The answer reward $R_{ans}$ is defined as:
\begin{equation}
    R_{ans}(y_{pred}, y_{gt}) = 
    \begin{cases} 
    1.0 & \text{if } y_{pred} = y_{gt} \\
    d(r(y_{pred}), r(y_{gt})) & \text{if } y_{pred} \neq y_{gt} \land y_{pred} \in \mathcal{Y} \\
    0 & \text{if } y_{pred} \notin \mathcal{Y}
    \end{cases}
\end{equation}
Here $d(\cdot)$ represents a semantic distance function that quantifies the proximity between the predicted rank and the ground truth rank. Furthermore, to ensure the model adheres to the structural constraints of the output format, we incorporate a format reward $R_{fmt}$. Consequently, the final training objective maximizes the cumulative reward defined as:
\begin{equation}
    R_{total} = \lambda_{ans}R_{ans} + \lambda_{fmt}R_{fmt}
\end{equation}
where $\lambda_{ans}$ and $\lambda_{fmt}$ are the balancing coefficients for the answer accuracy and format compliance, respectively.

\section{Experiments}
\label{sec:experiments}
\begin{table*}[h]
\centering
\caption{\textbf{Main Results on Three Adjudication Benchmarks.} We report Precision ($P$), Recall ($R$), and Accuracy ($Acc.$).Best results are in bold, second best are underlined.}
\vspace{-0.3pt}
\label{tab:main_results}
\resizebox{\textwidth}{!}{%
\scriptsize
\setlength{\tabcolsep}{3pt}
\renewcommand{\arraystretch}{1.15}
\begin{tabular}{cc|ccccc|ccccc|ccccc|c}
\toprule
\multirow{3}{*}{\textbf{Backbone}} & \multirow{3}{*}{\textbf{Method}} & \multicolumn{5}{c|}{\textbf{Appeal }} & \multicolumn{5}{c|}{\textbf{Driver-Cancel}} & \multicolumn{5}{c|}{\textbf{Passenger-Cancel}} & \multirow{3}{*}{\textbf{Overall}} \\
 & & \multicolumn{2}{c}{Normal} & \multicolumn{2}{c}{Malicious} & \multirow{2}{*}{$Acc.$} & \multicolumn{2}{c}{Normal} & \multicolumn{2}{c}{Malicious} & \multirow{2}{*}{$Acc.$} & \multicolumn{2}{c}{Normal} & \multicolumn{2}{c}{Malicious} & \multirow{2}{*}{$Acc.$} & \\
 & & $P$ & $R$ & $P$ & $R$ & & $P$ & $R$ & $P$ & $R$ & & $P$ & $R$ & $P$ & $R$ & & $Acc.$ \\
\midrule
\multicolumn{18}{c}{\cellcolor{refreshinggreen}\textit{\textbf{LLM Series}}} \\
\midrule
\multirow{2}{*}{DeepSeek-V3.1} & Standard & 70.91 & \textbf{95.49} & 80.70 & 20.63 & 57.89 & 80.69 & 86.67 & \textbf{64.00} & 84.21 & \underline{86.15} & 88.57 & 66.77 & 89.12 & 65.40 & 85.30 & 75.25 \\
 & CoT & 71.15 & 93.59 & 86.79 & 20.63 & 58.09 & 77.99 & \underline{91.85} & \underline{58.33} & 73.68 & \textbf{86.36} & 86.75 & 66.46 & 88.38 & 66.54 & 85.30 & 75.36 \\
\multirow{2}{*}{Qwen3-32B-Insruct} & Standard & 62.76 & 89.08 & 27.07 & 63.23 & 38.93 & 32.67 & 36.30 & 23.08 & 15.79 & 25.97 & 30.73 & 41.23 & 53.98 & 23.19 & 17.26 & 26.77 \\
 & CoT & 64.83 & 88.39 & 30.24 & 62.78 & 40.81& 39.55 & 39.26 & 42.86 & 31.58 & 19.91 & 30.33 & 39.38 & 53.66 & 16.73 & 16.72 & 26.21 \\
\multirow{2}{*}{QwQ-32B} & Standard & 72.77 & 75.04 & 58.60 & 48.88 & 60.28 & 38.19 & 40.74 & 25.00 & 21.05 & 26.62 & 32.47 & 46.46 & 53.91 & 23.57 & 19.44 & 35.82 \\
 & CoT & 76.88 & 77.82 & 66.66 & 36.77 & 61.57& 47.50 & 70.37 & 30.30 & 52.63 & 22.51 & 38.03 & 91.38 & 53.71 & 54.37 & 18.51 & 35.18 \\
\midrule
\multicolumn{18}{c}{\cellcolor{refreshinggreen}\textit{\textbf{MLLM Series}}} \\
\midrule
\multirow{2}{*}{Qwen3-VL-8B-Instruct} & Standard & 63.09 & 25.48 & 27.34 & 15.70 & 39.82 & \textbf{100.00} & 9.60 & 50.00 & 15.80 & 71.00 & \underline{92.90} & 8.00 & \textbf{92.30} & 9.10 & 73.60 & 60.61 \\
 & CoT & 65.57 & 24.09 & 51.35 & 8.52 & 41.81 & 90.00 & 13.30 & 54.50 & 31.60 & 71.60 & \textbf{95.70} & 13.80 & \underline{91.50} & 16.30 & 75.00 & 62.09 \\

\multirow{2}{*}{Qwen3-VL-8B-Thinking} & Standard & 75.00 & 4.16 & 0.89 & 77.78 & 2.78 & 50.00 & 2.20 & 0.00 & 0.00 & 13.90 & 66.70 & 1.20 & 33.30 & 0.40 & 8.70 & 7.37 \\
 & CoT & 91.30 & 3.64 & 100.00 & 5.83 & 3.48 & 100.00 & 0.70 & 0.00 & 0.00 & 13.00 & 0.00 & 0.00 & 0.00 & 0.00 & 11.60 & 8.82 \\
 
\multirow{2}{*}{Qwen3-VL-32B-Instruct} & Standard & 60.61 & 79.20 & 41.88 & 52.02 & 43.20 & \underline{93.50} & 43.00 & 45.80 & 57.90 & 78.40 & 84.60 & 37.20 & 81.50 & 38.40 & 78.90 & 65.55 \\
 & CoT & 60.57 & 83.36 & 38.29 & 16.14 & 41.71 & 91.00 & 52.60 & 53.60 & 78.90 & 72.10 & 83.60 & 50.20 & 81.30 & 52.90 & 68.50 & 59.14 \\

\multirow{2}{*}{Qwen3-VL-32B-Thinking} & Standard & 75.85 & 69.67 & 75.00 & 36.32 & 49.75 & 34.76 & 48.15 & 0.00 & 0.00 & 54.55 & 27.59 & 51.69 & 0.00 & 0.00 & 39.74 & 45.94 \\
 & CoT & 76.18 & 78.16 & 75.93 & 36.77 & 57.30 & 32.29 & 22.96 & 0.00 & 0.00 & 61.04 & 27.52 & 27.69 & 0.00 & 0.00 & 54.98 & 56.86 \\
 
\multirow{2}{*}{MiniCPM-V} & Standard & 67.44 & 5.03 & 0.00 & 0.00 & 14.90 & 53.33 & 17.78 & 0.00 & 0.00 & 33.12 & 36.84 & 4.31 & 66.67 & 1.52 & 18.66 & 19.68 \\
 & CoT & 70.18 & 6.93 & 0.00 & 0.00 & 16.48 & 44.00 & 8.15 & 0.00 & 0.00 & 16.88 & 35.85 & 5.85 & 45.45 & 1.90 & 8.71 & 12.96 \\
\midrule
\multicolumn{18}{c}{\cellcolor{refreshinggreen}\textit{\textbf{RideJudge}}} \\
\midrule
\multicolumn{2}{c|}{\textbf{RideJudge-4B}} & \underline{94.55} & 93.24 & \textbf{93.06} & \underline{90.13} & \underline{90.86} & 72.41 & \textbf{93.33} & 51.42 & \textbf{94.74} & 80.95 & 84.50 & \underline{85.54} & 87.69 & \textbf{86.69} & \underline{86.63} & \underline{87.25} \\
\multicolumn{2}{c|}{\textbf{RideJudge-8B}} & \textbf{94.60} & \underline{94.11} & \underline{91.52} & \textbf{91.93} & \textbf{91.86} & 78.85 & 91.11 & 41.86 & \underline{94.73} & 83.55& 89.76 & 80.92 & 86.48 & \underline{85.71} & \textbf{87.40} & \textbf{88.41} \\
\bottomrule
\end{tabular}%
}
\vspace{-0.2cm}
\end{table*}

\textbf{Baselines.}
We benchmark our model against three LLMs and five MLLMs to provide a comprehensive comparison. For rigorous evaluation, we employ both standard prompting and Chain-of-Thought prompting strategies across all baseline models. Note that the LLMs only receive the textual order information $O_{text}$ as input due to their inherent modality constraints.

\textbf{Implementation Details.}
We train two variants, RideJudge-8B and RideJudge-4B, which are initialized from Qwen3-VL-8B-Instruct and Qwen3-VL-4B-Instruct, respectively. The supervised fine-tuning stages are implemented using the LLaMA-Factory framework \cite{zheng2024llamafactory} while the reinforcement learning stage utilizes Easy-R1\cite{zheng2025easyr1}. All training experiments are conducted on 8 $\times$ NVIDIA H200 GPUs. Further hyperparameter details are provided in the Appendix \ref{app:training_details}.

\textbf{Evaluation Metrics.}
We primarily utilize Accuracy to measure the global correctness of the model across all liability types. To enable a more granular analysis regarding the ordinal level, we further categorize the fine-grained responsibility verdicts into two hierarchical levels named Normal and Malicious. The Normal category targets basic liability disputes to distinguish whether the driver is at fault or holds no responsibility. The Malicious category focuses on identifying severe violations where the driver exhibits intentionally malicious behaviors. For these specific binary classification tasks, we report the Precision and Recall to evaluate the sensitivity and exactness of the model in different contexts.
\subsection{Performance on Ride-Hailing Adjudications}

\label{sec:rq1_results}

\paragraph{Evaluation Benchmarks.}
    To validate the effectiveness of our framework in real-world settings, we conducted evaluations on datasets collected from DiDi Chuxing, which is one of the largest ride-hailing platforms globally. We selected three distinct test sets representing critical and complex adjudication scenarios.

 \textbf{Appeal:} This set contains 1,007 challenging samples where drivers formally appealed the initial platform verdict after order cancellation.

\textbf{Driver-Cancel:} This set comprises 453 selected hard samples involving reservation orders cancelled by drivers.

\textbf{Passenger-Cancel:} This set includes 1,249 selected hard samples involving reservation orders initiated by passengers.

Table \ref{tab:main_results} presents the comprehensive results across these three distinct benchmarks. Our framework secures the first or second best results across the majority of evaluation metrics, demonstrating robust generalisation capabilities. Specifically, our RideJudge-8B model achieves an overall accuracy of 88.41\% on the combined test set. This performance significantly surpasses the same series Qwen3-VL-32B-Instruct, which scores 65.55\%, and the reasoning enhanced text model DeepSeek-V3.1, which attains 75.25\%.
\begin{table}
\centering
\caption{\textbf{Main Results Comparison.} We report the overall Precision ($P$), Recall ($R$), and Accuracy ($Acc.$) of selected baselines and our method.}
\label{tab:simplified_results}
\small
\setlength{\tabcolsep}{10pt} 
\begin{tabular}{lccc}
\toprule
\textbf{Model} & \textbf{P} & \textbf{R} & \textbf{Acc.} \\
\midrule
QwQ-32B & 29.73 & 51.27 & 34.00\\
Qwen3-VL-8B-Instruct & 29.68 & 51.69 & 33.40 \\
Qwen3-VL-8B-Thinking & 29.67 & 52.77 & 33.90 \\
Qwen3-VL-32B-Instruct & 30.22 & 53.39 & 34.40 \\
Qwen3-VL-32B-Thinking & 30.12 & 53.88 & 33.40 \\
\midrule
\textbf{Ours} & \textbf{53.17} & \textbf{65.34} & \textbf{60.20} \\
\bottomrule
\end{tabular}
\vspace{-0.5cm}
\end{table}
A rigorous analysis of the baseline reveals a notable trend where text-only LLMs generally outperform standard MLLM baselines. This performance disparity stems from the fact that adjudication is inherently a logic-intensive task requiring strict adherence to complex platform rules. Text-only models like DeepSeek-V3.1 leverage their superior reasoning capabilities to infer verdicts based on textual metadata, whereas general-purpose MLLMs lack specific pre-training on domain-specific trajectory maps, leading to severe hallucinations. However, text-only models eventually hit a performance ceiling due to their inability to access the visual modality, which is essential for verifying spatial proofs. It is worth noting that certain baselines exhibit abnormally high precision in specific scenarios, such as Qwen3-VL-8B-Instruct, achieving 100.00\% precision on the Driver-Cancel Normal task. This metric is misleading as it is accompanied by a negligible recall of 9.60\% indicating that the model biases heavily towards negative predictions and fails to retrieve valid cases. Similarly, several general multimodal models fail to identify malicious intents entirely, yielding zero precision and recall in malicious categories, which stems from their lack of domain-specific alignment. RideJudge effectively bridges these gaps by integrating the logical rigor of LLMs with the precise visual grounding of trajectory data, resulting in superior performance across both normal and malicious adjudication tasks.
\subsection{Ablation and In-Depth Study}
\label{sec:ablation_study}
To dissect the contribution of each component in our framework, we conduct a comprehensive ablation study on the Appeal benchmark using the Qwen3-VL-8B backbone. As presented in Table \ref{tab:ablation}, we evaluate four variants by systematically removing key modules.

\begin{table}
  \caption{Ablation Study on Appeal Benchmark. $\Delta$ indicates the performance gap between the full method and the ablated version.}
  \label{tab:ablation}
  \centering
  \resizebox{\columnwidth}{!}{%
  \setlength{\tabcolsep}{3.5pt}
  \renewcommand{\arraystretch}{1.2}
  \begin{tabular}{c l cc cc c}
    \toprule
    \multirow{2}{*}{\textbf{No.}} & \multirow{2}{*}{\textbf{Settings}} & \multicolumn{2}{c}{\textbf{Normal}} & \multicolumn{2}{c}{\textbf{Malicious}} & \multirow{2}{*}{\textbf{Acc.}} \\
    \cmidrule(lr){3-4} \cmidrule(lr){5-6}
      & & $P$ & $R$ & $P$ & $R$ & \\
    \midrule
    
    1 & Baseline & 63.57 & 24.09 & 51.35 & 8.52 & 44.49 \\
    
    \hline\hline 
    
    2 & + 1st stage & 64.29 & 28.08 & 48.84 & 9.42 & 45.91 \\
    3 & + 2nd stage & 80.92 & 76.43 & 68.82 & 57.40 & 68.81 \\
    4 & + 3rd stage & 94.60 & 94.11 & 91.52 & 91.93 & 91.86 \\
    
    \hline\hline
    
    5 & SFT w/o CoA & 76.88 & 77.82 & 67.31 & 36.77 & 63.75 \\
    6 & SFT w/ CoA & 80.92 & 76.43 & 68.82 & 57.40 & 68.81 \\
    \rowcolor{deltagreen} 
     & $\Delta$ & \textcolor{textgreen}{\textbf{+4.04}} & \textcolor{textred}{\textbf{-1.39}} & \textcolor{textgreen}{\textbf{+1.51}} & \textcolor{textgreen}{\textbf{+20.63}} & \textcolor{textgreen}{\textbf{+5.06}} \\
    
    \midrule 
    
    7 & w/o KACR & 84.17 & 78.34 & 70.05 & 68.16 & 70.56 \\
    8 & w/ KACR & 94.60 & 94.11 & 91.52 & 91.93 & 91.86 \\
    \rowcolor{deltagreen} 
     & $\Delta$ & \textcolor{textgreen}{\textbf{+10.43}} & \textcolor{textgreen}{\textbf{+15.77}} & \textcolor{textgreen}{\textbf{+21.47}} & \textcolor{textgreen}{\textbf{+23.77}} & \textcolor{textgreen}{\textbf{+21.30}} \\
    
    \midrule 
    
    9 & DAPO w/o OS Reward & 85.97 & 66.90 & 72.83 & 56.50 & 70.51 \\
    10 & DAPO w/ OS Reward & 94.60 & 94.11 & 91.52 & 91.93 & 91.86 \\
    \rowcolor{deltagreen} 
     & $\Delta$ & \textcolor{textgreen}{\textbf{+8.63}} & \textcolor{textgreen}{\textbf{+27.21}} & \textcolor{textgreen}{\textbf{+18.69}} & \textcolor{textgreen}{\textbf{+35.43}} & \textcolor{textgreen}{\textbf{+21.35}} \\
    
    \bottomrule
  \end{tabular}%
  }
  \vspace{-10pt}
\end{table}
\textbf{Effect of Training Strategy.}We first examine the progressive impact of our training paradigm. The transition from the Baseline to the 1st stage yields a marginal improvement of 1.42\%. This limited gain suggests that while Visual Concept Alignment successfully grounds geometric patterns, it fails to verify complex liability scenarios due to the absence of reasoning capabilities. In contrast, the introduction of the 2nd stage brings a substantial performance leap, raising the accuracy to 68.81\%. This confirms that the Juridical Logic Alignment derived from our Chain of Adjudication effectively bridges the gap between perception and decision-making. Finally, the 3rd stage involving Reinforcement Learning Tuning propels the model to its peak performance of 91.86\% demonstrating that exploring decision boundaries via Group Relative Policy Optimization significantly consolidates the robustness of the model.

\textbf{Impact of Key Modules.}We further investigate the specific contributions of our architectural designs. Comparing No.5 and No.6 reveals that integrating the structured Chain of Adjudication outperforms standard unstructured supervision. The CoA framework improves the Recall on Malicious tasks by over 20\%, indicating that decomposed reasoning steps help the model uncover subtle fraud patterns that end-to-end learning overlooks. Furthermore, the comparison between No.9 and No.10 highlights the criticality of our Ordinal Sensitive Reward. Removing this mechanism and relying solely on standard binary rewards causes a drastic accuracy drop of 21.35\%. This sharp decline verifies that penalizing logical deviations based on their ordinal level is essential for stabilizing the reinforcement learning process in high-stakes adjudication tasks. Similarly, the removal of the KACR module in No.7 leads to a significant performance regression, which further validates the necessity of rigorous knowledge alignment and data selection in ensuring high-confidence predictions.

\begin{figure*}[h]
    \centering
    \includegraphics[width=1\textwidth, trim=0.1cm 2.8cm 0.1cm 2.8cm, clip]{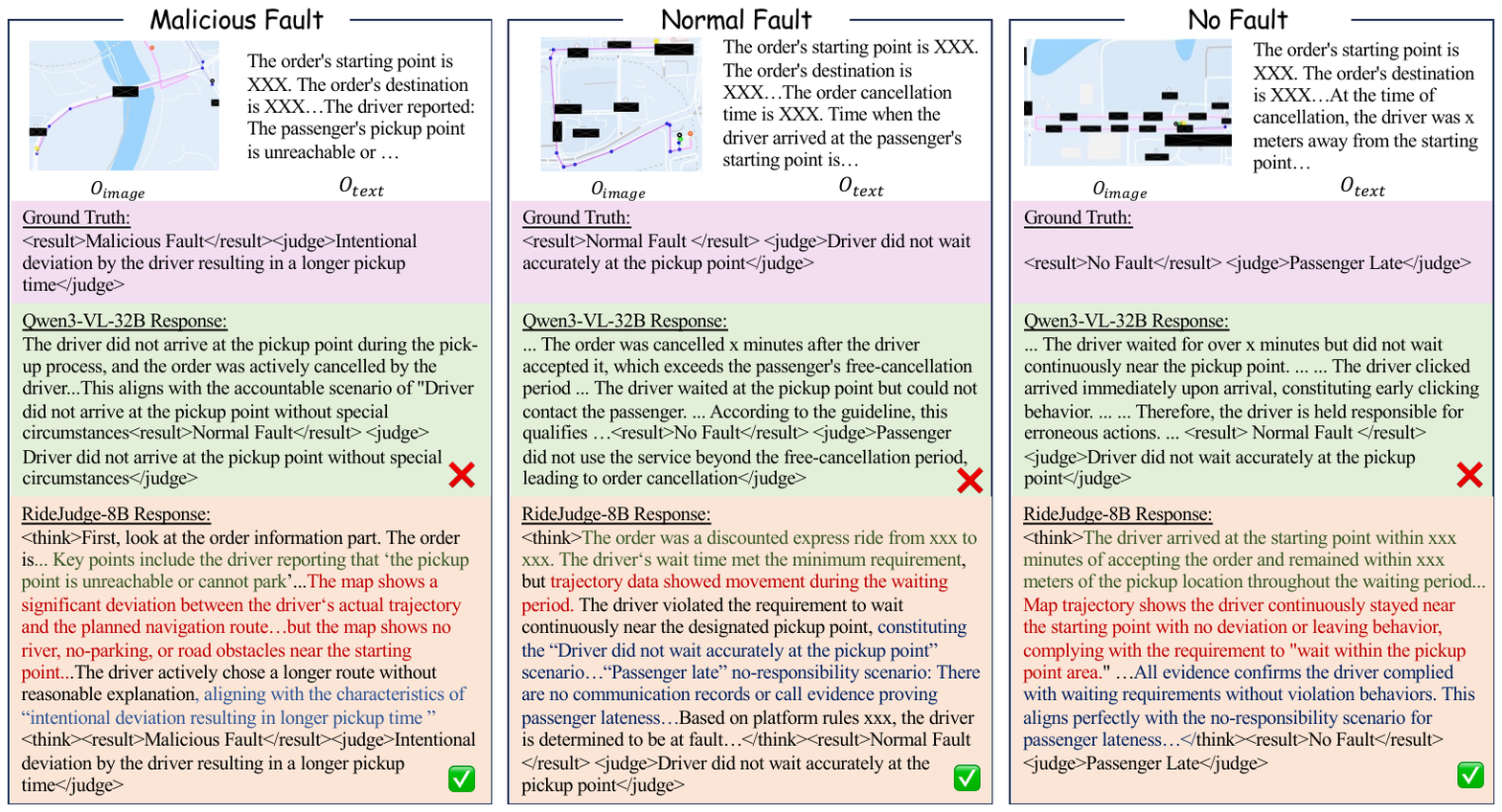}
\caption{Qualitative case studies on the Appeal benchmark. To preserve privacy, sensitive textual regions in the images, as well as specific numerical values and location names within the reasoning chains, have been masked. We highlight the \textcolor{green}{Information Analysis} process in green, the \textcolor{red}{Visual Evidence Integration} process in red, and the \textcolor{blue}{Rule Grounding} process in blue.}
    \label{fig:case}
\end{figure*}
\textbf{Scaling and Stability.}To assess the scalability of our synthesized data, we trained the Stage 2 model using 30\%, 50\%, and 70\% subsets of the CoA dataset. As shown in Figure \ref{fig:a+b} left, model performance steadily improves as data volume increases, confirming the high quality and efficacy of our synthesis pipeline. Furthermore, to verify robustness in real-world scenarios, we sampled 5,000 real appeal orders from ten major Chinese cities. Figure \ref{fig:a+b} right illustrates that RideJudge-8B maintains consistent high accuracy across diverse urban environments, demonstrating strong geographical generalization and stability.

\textbf{Case Study.}Figure \ref{fig:case} visualizes the reasoning  of RideJudge-8B compared to Qwen3-VL-32B-Instruct. Our model sequentially performs Information Analysis (green), Visual Evidence Integration (red), and Rule Grounding (blue). This structured reasoning allows the model to accurately associate the case with specific rules and derive the correct verdict.

\begin{figure}
        \label{fig:a+b}
    \centering
    \begin{subfigure}[c]{0.40\linewidth}
        \centering
        \includegraphics[width=\linewidth]{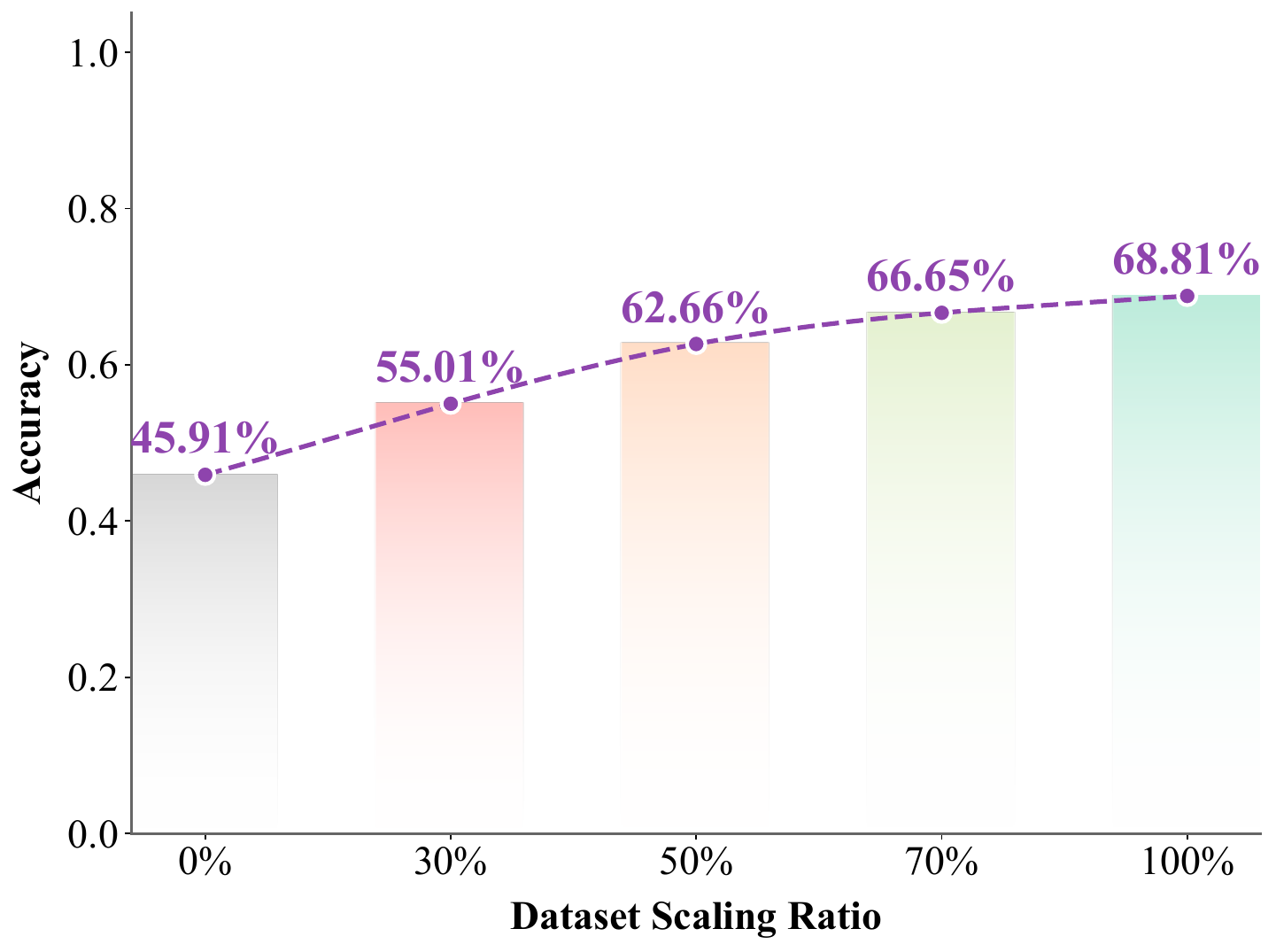}
        \label{fig:temporal_bar}
    \end{subfigure}
    \hfill 
    \begin{subfigure}[c]{0.38\linewidth}
        \centering
        \includegraphics[width=\linewidth]{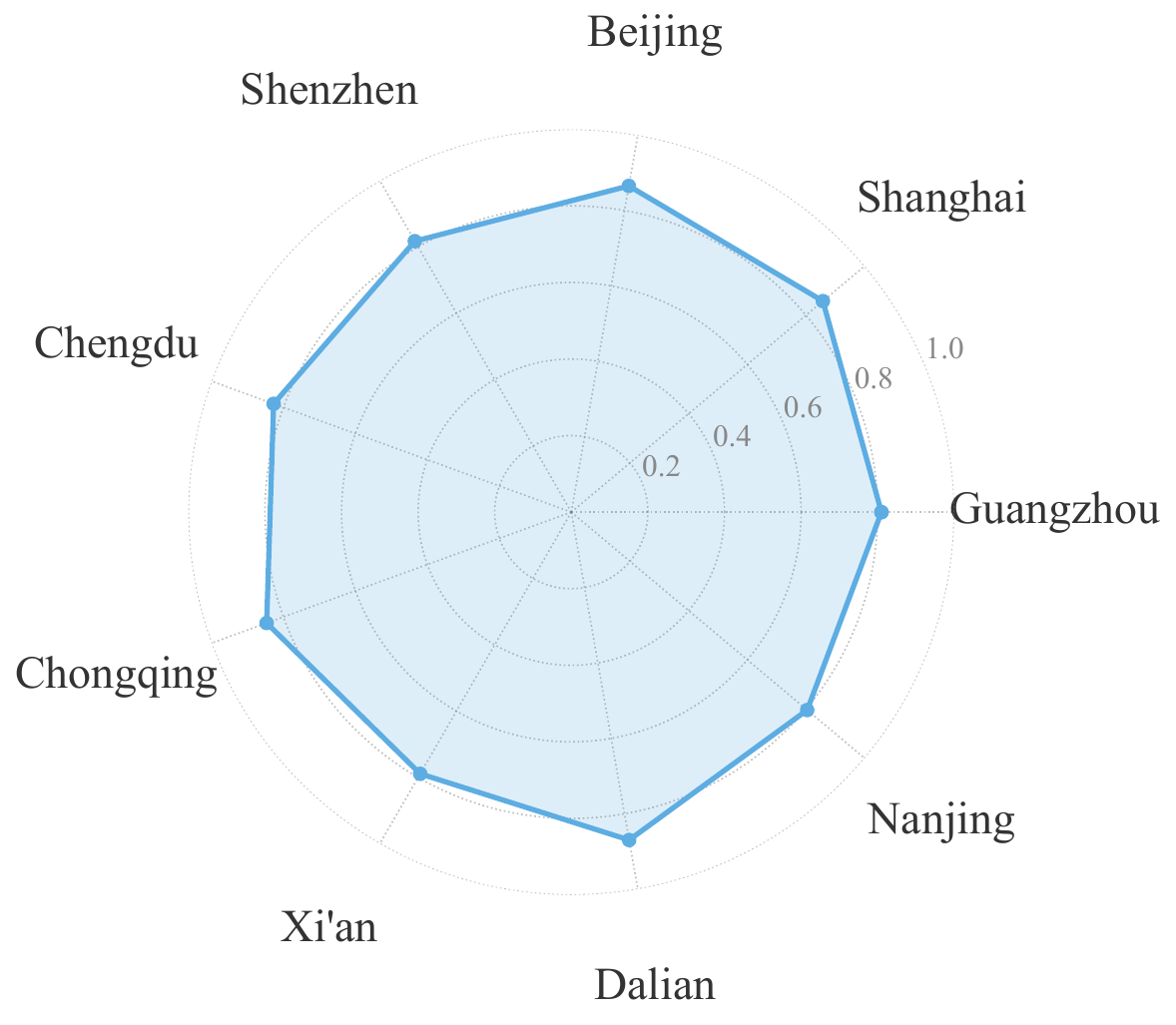}
        \label{fig:spatial_radar}
    \end{subfigure}
    \caption{\textbf{Left:} Performance scaling with our CoA synthetic data used in Stage 2. \textbf{Right:} Stability analysis of RideJudge-8B from ten major cities.}
    \vspace{-15pt}
    \label{fig:robustness_analysis}
            \label{fig:a+b}

\end{figure}

\subsection{Performance on Other Multimodal Adjudication Tasks}

To demonstrate the generalization of our framework, we extend our evaluation to the PetFinder Prediction benchmark. We contend that this task represents a generalized form that shares an intrinsic similarity with dispute adjudication: both require synthesizing heterogeneous evidence, comprising unstructured textual descriptions, structured tabular metadata (e.g., age, health), and visual imagery, to perform a judgment against a backdrop of implicit valuation rules and historical precedents.The objective is to predict the speed of adoption classified into discrete categories.

As presented in Table~\ref{tab:simplified_results}, our trained model demonstrates robust performance that significantly outperforms standard multimodal baselines across all metrics. The results indicate that our core mechanisms, specifically the reasoning synthesis and dynamix context , are not confined to trajectory data but effectively generalize to other heterogeneous data structures. This empirical evidence underscores the versatility and strong transferability of our Visual-Logic-Aligned architecture in broader multimodal decision-making scenarios. For more comprehensive implementation details and experimental settings, please refer to Appendix~\ref{app:petfinder_details}.

\section{Related Work}
\subsection{Large Language Models for Automated Adjudication}
The application of  LLMs in the legal domain has transitioned from generic text processing to specialized juridical reasoning frameworks. Early domain-specific models, such as ChatLaw~\cite{chatlaw} and LawGPT~\cite{lawgpt}, primarily focused on injecting legal vocabulary and retrieving statutory knowledge to handle complex queries. To enhance logical robustness, recent frameworks like SaulLM~\cite{saullm} and DISC-LawLLM~\cite{disc-lawllm} have integrated prompting strategies to simulate the procedural reasoning of human judges. Furthermore, the ``LLM-as-a-Judge'' paradigm~\cite{judgelm,prometheus} has formalized the use of models for impartial evaluation and scoring. However, a critical limitation persists across these works: they operate solely on textual precedents. In real-world ride-hailing adjudication, liability assessment fundamentally relies on physical evidence, specifically the alignment between spatial-temporal trajectories and map semantics. Existing legal LLMs lack the visual perception capabilities required to ground abstract liability regulations into concrete trajectory patterns, a gap our framework addresses through a progressive visual-logic alignment mechanism.

\subsection{Multimodal Intelligence in Spatio-Temporal Decision Making}
Our work also intersects with the  field of MLLMs for urban and autonomous driving applications. Recent research has demonstrated the efficacy of integrating spatiotemporal dependencies into transformer architectures. For instance, UrbanGPT~\cite{urbangpt}  models city-wide flow dynamics for traffic prediction, while TrafficGPT~\cite{trafficgpt} explores LLMs as control agents for traffic signal optimization. In the context of visual understanding for driving, NuScenes-QA~\cite{qian2024nuscenesqa} and DriveLM~\cite{sima2024drivelm} have established benchmarks for Graph Visual Question Answering. Despite these advancements, existing models focus primarily on \textit{predictive} tasks or \textit{perceptual} Q\&A . They lack the specific capability to perform \textit{forensic} analysis, which requires auditing historical behaviors against strict regulatory frameworks to attribute liability.

\section{Conclusion}
In this paper, we presented RideJudge, a Progressive Visual-Logic-Aligned Framework for the ride-hailing adjudication. By integrating automated data synthesis, adaptive context optimization, and a three-stage training paradigm, our approach effectively bridges the systemic disconnects between general MLLMs and rigorous adjudication requirements. Extensive results on real-world tasks demonstrate the effectiveness of the proposal. This work offers a scalable blueprint for deploying specialized MLLMs in broader, complex, rule-governed decision-making tasks.

\begin{acks}
This work is supported by the Key Program of the Jiangsu Science Foundation (Grant No. BK20243012), the National Natural Science Foundation of China (Grant No. 62306133), and the CCF-DiDi GAIA Collaborative Research Funds for Young Scholars. We also appreciate the guidance provided by the engineers in Didichuxing Co. Ltd.
\end{acks}
\clearpage
\bibliographystyle{ACM-Reference-Format}
\bibliography{sample-base}
\appendix
\clearpage
\section{Ethical Statement}

To promote reproducibility and facilitate further research in automated adjudication we are committed to open sourcing the complete code base for our proposed methods. Additionally we will publicly release\textbf{ SynTra}j which serves as a high-quality synthetic trajectory dataset. 

Regarding the proprietary ride-hailing data used in this work we strictly adhere to ethical guidelines regarding data privacy and user protection throughout this study. All adjudication data utilized in our experiments has undergone a rigorous de-identification process to ensure that no Personally Identifiable Information of drivers or passengers is exposed. The usage of this data is strictly restricted to the research and development of platform adjudication algorithms. Furthermore all data collection was conducted in full compliance with the platform data governance regulations and explicit informed consent regarding data usage for service improvement was obtained from all involved parties prior to acquisition. Consequently all model evaluations were performed using local deployments to ensure no sensitive data was transmitted to external third-party APIs. 

Due to these strict privacy protocols we are unable to release the real-world order dataset at this time. However the extensive evaluation performed on the publicly available SynTraj dataset effectively validates the generalization capability and robustness of our approach.
\section{Details of Methods}

\subsection{SynTraj Mutation}
\label{app:mutation_protocols}

In this section, we provide the formal construction protocols for the liability samples in SynTraj. We denote the base compliant geometric plan as $S_{geo} = [p_1, \dots, p_N]$, where $p_1=l_s$ (start) and $p_N=l_e$ (end). We define the navigation oracle $\Psi(x, y)$ as a function returning the coordinate sequence of the shortest path connecting location $x$ to $y$ via the road network.

\subsubsection{Trajectory Drift}
To simulate realistic sensor imperfections and GPS jitter, we apply a stochastic perturbation mechanism. Unlike topological violations, this mutation operates directly on the coordinate level without altering the route logic.
\begin{itemize}
    \item \textbf{Noise Injection:} For a trajectory sequence $\mathcal{T} = [q_1, \dots, q_T]$, we generate the drifted observation $\mathcal{T}_{drift} = [q'_1, \dots, q'_T]$, where each point is perturbed independently:
    \begin{equation}
        q'_t = q_t + \vec{\epsilon}_t, \quad \vec{\epsilon}_t \sim \mathcal{N}(0, \sigma^2 \mathbf{I})
    \end{equation}
    We set $\sigma \approx 10\text{-}15m$ to simulate urban canyon effects. This forces the model to learn robust geometric alignment features rather than overfitting to perfectly smooth synthetic lines.
\end{itemize}

\subsubsection{Unintentional Deviation}
This mutation simulates a driver diverging from the planned route at an intermediate intersection, potentially to take a shortcut or due to a navigation error, but eventually returning to the destination.
\begin{itemize}
    \item \textbf{Step I: Anchor \& Waypoint Generation.} We sample a split node $p_t \in S_{geo}$ ($1 < t < N$) and compute the intended heading vector $\vec{v} = p_{t+1} - p_t$. We generate an off-route waypoint $w_{dev}$ via rotation:
    \begin{equation}
        w_{dev} = p_t + \lambda \cdot \mathcal{R}_{\theta}(\vec{v}), \quad \theta \in \{90^\circ, 270^\circ\}
    \end{equation}
    where $\mathcal{R}_{\theta}$ is the rotation matrix and $\lambda$ is the deviation magnitude.
    
    \item \textbf{Step II: Re-routing and Stitching.} We query the oracle $\Psi$ to bridge the topological gap. The final trajectory $\mathcal{T}_{yaw}$ is synthesized by stitching the compliant history, the deviation path, and the recovery path:
    \begin{equation}
        \mathcal{T}_{yaw} = \underbrace{S_{geo}[1:t]}_{\text{History}} \oplus \underbrace{\Psi(p_t, w_{dev})}_{\text{Deviation}} \oplus \underbrace{\Psi(w_{dev}, l_e)}_{\text{Recovery}}
    \end{equation}
\end{itemize}

\subsubsection{Reverse Driving}
To synthesize high-risk retrograde motion, we construct a trajectory that visually overlaps with the planned route but flows against the topological direction of the lane.
\begin{itemize}
    \item \textbf{Step I: Vector Inversion.} At a selected critical node $p_t$, we calculate the intended heading $\vec{v}$ and derive a "wrong-way" target $w_{rev}$ using an obtuse rotation angle:
    \begin{equation}
        w_{rev} = p_t + \lambda \cdot \mathcal{R}_{\phi}(\vec{v}), \quad \phi \in [150^\circ, 210^\circ]
    \end{equation}
    
    \item \textbf{Step II: Truncated Synthesis.} We generate the violation segment targeting $w_{rev}$ but truncate it to a distance limit $\delta$ to simulate the onset of the violation. The final trajectory is defined as:
    \begin{equation}
        \mathcal{T}_{rev} = \underbrace{S_{geo}[1:t]}_{\text{History}} \oplus \underbrace{\text{Trunc}(\Psi(p_t, w_{rev}), \delta)}_{\text{Reverse Segment}}
    \end{equation}
    This creates a sharp semantic conflict where the visual motion opposes the map's permitted lane directionality.
\end{itemize}

\subsubsection{Arrival-then-Leave}
This scenario models the "refusal to operate" behavior where a driver completes the order but subsequently departs from the pickup location $l_e$ instead of waiting.
\begin{itemize}
    \item \textbf{Step I: Escape Target Generation.} Upon validating the full execution of the geometric plan $S_{geo}$, we generate an escape target $w_{esc}$ located at a Euclidean distance $d > \tau_{thresh}$ from $l_e$ with a random bearing.
    
    \item \textbf{Step II: Sequence Extension.} The final violation trajectory is constructed as the concatenation of the arrival phase and the unauthorized departure phase:
    \begin{equation}
        \mathcal{T}_{leave} = \underbrace{S_{geo}}_{\text{Arrival}} \oplus \underbrace{\Psi(l_e, w_{esc})}_{\text{Departure}}
    \end{equation}
    This introduces a multimodal mismatch: the instruction sequence $S_{inst}$ terminates at $l_e$ (implying a stop), while the visual trajectory $\mathcal{T}_{leave}$ continues to evolve spatially.
\end{itemize}
\subsection{Chain-of-Adjudication}
\label{sec:coa_details}

In our Chain-of-Adjudication framework, we assign specific model architectures to each role to maximize performance while maintaining computational efficiency. Both the Adjudicator and the Reasoning Refiner are instantiated using the QwQ-32B model leveraging its advanced logical reasoning capabilities for rule interpretation and text summarization. The Visual Analyst is powered by Qwen3-VL-32B-Instruct enabling precise interpretation of spatiotemporal trajectory data. To ensure the reasoning process remains focused and to prevent infinite recursive loops during the inquiry phase we strictly limit the maximum number of interaction turns between the Adjudicator and the Visual Analyst to 8 turns.

Below we provide the specific system prompts designed for each agent in the framework.

\begin{tcolorbox}[
    colback=black!5!white,
    colframe=black,
    coltitle=white,
    fonttitle=\bfseries\ttfamily,
    title=The prompt for the Adjudicator,
    halign title=center,
    sharp corners=all,
    rounded corners=downhill,
    arc=3mm,
    boxrule=0.8pt,
    borderline={0.4pt}{0pt}{dashed},
    before skip=6pt,  
    after skip=6pt,   
]
\label{prompt-adjudicator}
You are a ride-hailing marketplace adjudication expert responsible for determining which liability features and final liability a driver's behavior matches in a cancelled order based on the information provided after the driver accepts the request. Please provide the reasoning process and final result according to the adjudication rules and order information below. During the adjudication process, if you believe map-related information is necessary, you may use <map>map-related question</map> to ask. Please note that your final liability determination must be consistent with the human-annotated result.
\par
\end{tcolorbox}

\begin{tcolorbox}[
    colback=black!5!white,
    colframe=black,
    coltitle=white,
    fonttitle=\bfseries\ttfamily,
    title=The prompt for the Visual Analyst,
    halign title=center,
    sharp corners=all,
    rounded corners=downhill,
    arc=3mm,
    boxrule=0.8pt,
    borderline={0.4pt}{0pt}{dashed},
    before skip=6pt,  
    after skip=6pt,   
]
\label{prompt-visual-analyst}
You are a Map Expert responsible for liability adjudication in the ride-hailing marketplace. Your task is to analyze orders cancelled by drivers after acceptance and answer questions from an Adjudication Expert to determine which specific fault indicators the driver's behavior matches and the final liability. We will provide the order details, the map, and the expert's questions. Please carefully analyze the order information in the context of the map and return your answer within <answer></answer> tags.
\par
\end{tcolorbox}

\subsection{Knowledge-Aware Context Refinement}
\label{sec:knowledge_refinement_details}

In this section we elaborate on the training protocols and hyperparameter configurations for the Context Refinement module.

\subsubsection{Scenario-Aware Rule Calibration Details}

The Decomposed Ensemble Calibrator is designed to handle the high-dimensional and imbalanced nature of rule applicability. For each rule $r_i \in \mathcal{K}$ we train a dedicated binary classifier $f_i$.

\textbf{Hybrid Feature Representation.}
To capture both the structured metadata and the unstructured semantic context of an order we construct a hybrid feature space.
\begin{itemize}
    \item \textbf{Tabular Features:} We utilize intrinsic order attributes such as time of day, location coordinates, and cancellation reason codes as dense numerical features.
    \item \textbf{Semantic Features:} We concatenate textual fields including passenger complaints and driver appeals and encode them into 1024-dimensional embeddings using the pre-trained \texttt{bge-large-zh-v1.5} model.
\end{itemize}
These features are concatenated to form the input vector for the classifiers.
\begin{tcolorbox}[
    colback=black!5!white,
    colframe=black,
    coltitle=white,
    fonttitle=\bfseries\ttfamily,
    title=The prompt for the Reasoning Refiner,
    halign title=center,
    sharp corners=all,
    rounded corners=downhill,
    arc=3mm,
    boxrule=0.8pt,
    borderline={0.4pt}{0pt}{dashed},
    before skip=6pt,  
    after skip=6pt,   
]
\label{prompt-refiner}
You are a Reasoning Refinement Specialist in the ride-hailing adjudication domain. Your task is to act as a meta-cognitive editor to distill the raw, fragmented interaction history between an Adjudicator and a Visual Analyst into a coherent, standardized adjudication log.

Based on the provided conversation history, order metadata, and adjudication rules, please reconstruct the reasoning process and output the final conclusions following this strict format:

1. \textbf{Reasoning Chain}: Enclose your detailed reconstruction of the adjudication path within <reason>...</reason> tags. Inside this tag, you must structure the content into four distinct stages:
    (1) Information Analysis: Systematically summarize the key order metadata and dispute context.
    (2) Visual Evidence Integration: Synthesize the objective trajectory facts verified by the Visual Analyst.
    (3) Rule Grounding: Explicitly map the established facts to the specific liability clauses.
    (4) Comprehensive Adjudication: Perform the final logical deduction.

2. \textbf{Scenario Identification}: Output the specific adjudication scenario or fault category within <judge>...</judge> tags.

3. \textbf{Final Verdict}: Output the final liability determination within <result>...</result> tags.

Please ensure that the refined reasoning path is logically fluid and that the final determination aligns strictly with the provided ground truth.
\par
\end{tcolorbox}
\textbf{Adaptive Model Selection Strategy.}
Recognizing that different rules exhibit varying statistical distributions we do not rely on a single algorithm. Instead we implement an automated model search framework. For each rule $r_i$ we train three gradient boosting variants: XGBoost, LightGBM, and CatBoost.

During the validation phase we prioritize the Recall metric over Accuracy. This is a critical design choice for adjudication as missing an applicable rule is significantly more detrimental than retrieving a marginally relevant one. Consequently for every rule $r_i$ the final classifier $f_i$ is selected as follows:
\begin{equation}
    f_i = \mathop{\arg\max}_{m \in \{\text{XGB, LGB, Cat}\}} \text{Recall}_{val}(m)
\end{equation}
To address the extreme class imbalance where rule applicability is often a rare event we employ a dynamic down-sampling strategy on the majority class during training ensuring a balanced distribution for gradient optimization.

\subsubsection{Retrieval-Augmented Insight Extraction}

\textbf{Strict Temporal Partitioning.}
To ensure the rigorous evaluation of the system's predictive capability, we enforce a strict temporal cutoff for the historical repository $\mathcal{D}_{hist}$. When processing a query order at timestamp $t_{query}$, the repository is restricted to:
\begin{equation}
    \mathcal{D}_{hist}^{valid} = \{ (O^{(j)}, y^{(j)}) \mid t_{order}^{(j)} < t_{query} \}
\end{equation}
This prevents any form of data leakage where future adjudication outcomes could influence current decisions.

\textbf{Hyperparameters and Model Configuration.}
Based on empirical ablation studies, we set the retrieval depth to $K=4$, which offers the optimal trade-off between context richness and input noise. The Summary Agent responsible for synthesizing the Meta-Insight $I_{syn}$ is instantiated using QwQ-32B. Consistent with the calibration module, we employ the \texttt{bge-large-zh-v1.5} encoder for vectorizing historical precedents ensuring semantic space alignment.

\section{Training Details}
\label{app:training_details}

In this section, we provide the detailed hyperparameter configurations and training protocols for our three-stage Progressive Alignment Framework.

\textbf{Overview.}
 The supervised fine-tuning stages are implemented using the LLaMA-Factory framework while the reinforcement learning stage utilizes Easy-R1. All training experiments are conducted on a cluster of 8 NVIDIA H200 GPUs.

\subsection{Supervised Fine-Tuning Stages}

The first two stages focus on concept grounding and logic alignment through supervised learning. The specific hyperparameter settings are compared in Table~\ref{tab:sft_hyperparams}.

\textbf{Stage 1: Visual Concept Alignment.}
We utilize the SynTraj dataset to align visual representations with ride-hailing concepts. Consistent with our goal of grounding abstract rules into visual patterns, we unfreeze both the vision tower and the multimodal projector while keeping the language model frozen. This ensures the visual encoder adapts to the synthetic trajectory domain without altering the pre-trained linguistic knowledge.

\textbf{Stage 2: Adjudication Logic Alignment.}
We subsequently freeze the vision tower to preserve the learned visual features and perform full-parameter fine-tuning on the language model backbone. This stage aligns the reasoning chains of the model with professional adjudication protocols.

\begin{table}[h]
    \centering
    \caption{Hyperparameter settings for Stage 1 Visual Alignment and Stage 2 Logic Alignment.}
    \label{tab:sft_hyperparams}
    \small
    \setlength{\tabcolsep}{10pt}
    \begin{tabular}{lcc}
        \toprule
        \textbf{Hyperparameter} & \textbf{Stage 1} & \textbf{Stage 2} \\
        \midrule
        Vision Tower & Unfrozen & Frozen \\
        Projector & Unfrozen & Frozen \\
        Language Model & Frozen & Unfrozen \\
        \midrule
        Precision & BF16 & BF16 \\
        Optimizer & AdamW & AdamW \\
        Learning Rate & $5.0 \times 10^{-6}$ & $5.0 \times 10^{-6}$ \\
        LR Scheduler & Cosine & Cosine \\
        Warmup Ratio & 0.1 & 0.1 \\
        Per-Device Batch Size & 16 & 2 \\
        Gradient Accumulation & 2 & 2 \\
        Num Epochs & 4 & 8 \\
        \bottomrule
    \end{tabular}
\end{table}

\subsection{Reinforcement Learning Stage}

In the final stage, we employ the DAPO algorithm to explore decision boundaries and enhance robustness. We initialize the policy model from the Stage 2 checkpoint. DAPO is specifically selected for its ability to handle group-wise optimizations without requiring a separate value model making it highly efficient for reasoning tasks. The hyperparameters are detailed in Table~\ref{tab:rl_hyperparams}.The total reward is defined as the weighted sum of our OS answer reward and the format compliance reward with balancing coefficients 0.8 and 0.2 respectively. The format reward enforces structural compliance.
\begin{table}[h]
    \centering
    \caption{Hyperparameter settings for Stage 3 Reinforcement Learning via DAPO.}
    \label{tab:rl_hyperparams}
    \small
    \begin{tabular}{lc}
        \toprule
        \textbf{Hyperparameter} & \textbf{Value} \\
        \midrule
        Algorithm & DAPO \\
        Actor Learning Rate & $1.0 \times 10^{-6}$ \\
        Critic Learning Rate & $1.0 \times 10^{-6}$ \\
        Weight Decay & $1.0 \times 10^{-2}$ \\
        KL Coefficient & $0.01$ \\
        KL Penalty Type & Low Var KL \\
        \midrule
        Global Batch Size & 128 \\
        Rollout N & 5 \\
        Max Grad Norm & 1.0 \\
        Temperature / Top-p & 1.0 / 1.0 \\
        \bottomrule
    \end{tabular}
\end{table}

\section{Experimental Details on Multimodal Adjudication Tasks}
\label{app:petfinder_details}

To evaluate the generalization capability of our framework, we conducted experiments on the PetFinder Adoption Prediction benchmark. This section provides a formal definition of the task, details the data partition strategy used to simulate a retrieval-augmented adjudication scenario, and describes the construction of the domain-specific rule base.

\subsection{Task Definition}
The experiment utilizes the public dataset from the PetFinder.my Adoption Prediction challenge. We formulate the problem as a multimodal ordinal regression task involving heterogeneous input modalities. Let $\mathcal{X}$ denote the input space consisting of tuples $(I, T, M)$, where $I$ represents visual data from pet images, $T$ denotes unstructured textual descriptions provided by rescuers, and $M$ represents structured tabular metadata including attributes such as age, breed, gender, and health condition.

The objective is to predict a target label $y \in \{0, 1, 2, 3, 4\}$, which represents the adoption speed category. The labels correspond to ordinal time intervals ranging from same-day adoption to cases where the pet remains unadopted after 100 days. Unlike traditional classification approaches, we treat this as an adjudication task where the model must synthesize evidence from visual and textual modalities to determine the adoptability score.

To enable a rigorous evaluation of the model's discriminative capability, we further define a \textbf{binary classification sub-task}. Specifically, we aggregate the ordinal labels into two distinct categories: labels $\{0, 1, 2\}$ are remapped as \textit{Positive Samples} (representing short adoption times), while labels $\{3, 4\}$ are remapped as \textit{Negative Samples} (indicating prolonged stays or unadopted cases). This binarization allows us to report binary Precision and Recall metrics, thereby quantifying the model's effectiveness in distinguishing high-adoptability pets from those facing adoption challenges.

\subsection{Dataset Construction}
The original dataset contains 14,993 samples. To align with the case-based reasoning mechanism proposed in our framework, we restructured the dataset into three distinct subsets based on index ordering. This approach prevents information leakage and simulates a realistic scenario where past cases are used to adjudicate current inquiries.

We partitioned the data as follows:
\begin{enumerate}
    \item \textbf{Historical Case Library}: We designated the first 3,000 samples as the historical repository. This subset serves as the retrieval corpus $\mathcal{D}_{retrieval}$ for the Retrieval-Augmented Generation module. During inference, the model retrieves relevant cases from this fixed library to support its decision-making process.
    \item \textbf{Training Set}: The subsequent 10,000 samples, specifically indices 3,000 through 13,000, were utilized as the training set $\mathcal{D}_{train}$. This subset is used to optimize the parameters of the multimodal encoder and the reasoning policy.
    \item \textbf{Test Set}: The final 1,000 samples were reserved strictly for evaluation. This test set $\mathcal{D}_{test}$ contains the target cases for adjudication, ensuring that the performance metrics reflect the model's ability to generalize to unseen profiles using the established historical library.
\end{enumerate}

\subsection{Construction of Rule Knowledge Base}
A core component of our architecture is the integration of explicit domain knowledge to guide the reasoning process. We constructed a specialized rule base containing 48 distinct heuristics governing pet adoption adjudication.

The rule generation process employed a collaborative approach combining human domain expertise with Large Language Models. First, human experts identified fundamental biological and medical constraints derived from animal welfare guidelines. These core constraints formed the initial seed set. Second, we utilized a Large Language Model to analyze high-confidence samples from the Historical Case Library. The model extracted latent correlations between multimodal features and adoption speeds. These candidate rules were then formalized into logical implications. Finally, the generated rules underwent a manual verification process to eliminate redundancy and ensure logical consistency. The resulting set of 48 rules was compiled into the final knowledge base used during the cross-modal reasoning stage.
\end{document}